\documentclass[letterpaper, 10 pt, conference]{ieeeconf}  % Comment this line out if you need a4paper

\IEEEoverridecommandlockouts                              % This command is only needed if 
\newcommand{\RNum}[1]{\uppercase\expandafter{\romannumeral #1\relax}}
\usepackage{graphics} % for pdf, bitmapped graphics files
\usepackage{graphicx}
\graphicspath{{./figure/}}
\usepackage{subcaption}
\usepackage[font=small]{caption}

\usepackage{bm}

\usepackage[noadjust]{cite}

\usepackage{hyperref}
\usepackage{amsmath} % assumes amsmath package installed
\usepackage{amssymb}  % assumes amsmath package installed
\usepackage{algorithm}
\usepackage{algorithmic}

\usepackage{booktabs}
\usepackage{color}
\usepackage{xcolor}
\usepackage{enumerate}
\usepackage{svg}
\title{\LARGE \bf
Origami-based Shape Morphing Fingertip to Enhance Grasping Stability and Dexterity
}

\author{Zicheng Kan$^{1\dag}$, Yazhan Zhang$^{1\dag}$, \textit{Student Member, IEEE},  Chohei Pang$^{1}$,  \\
	and Michael Yu Wang$^{2}$, \textit{Fellow, IEEE}% <-this % stops a space
	\thanks{*Research is supported by the Hong Kong Innovation and Technology
Fund (ITF) ITS-018-17FP.}% <-this % stops a space
	\thanks{$^{1}$Z. Kan, Y. Zhang and C. Pang are with the Department of
Mechanical and Aerospace Engineering, Hong Kong University of Science
and Technology, Hong Kong (e-mail: zkan@connect.ust.hk; yzhangfr@connect.ust.hk; chpangad@conect.ust.hk).}% (PhD Candidate) (PhD Candidate)
	\thanks{$^{2}$M. Y. Wang (corresponding author) is with the Department of Mechanical
and Aerospace Engineering and the Department of Electronic and Computer
Engineering, Hong Kong University of Science and Technology, Hong Kong
(tel.: +852-34692544; e-mail: mywang@ust.hk).}%
    \thanks{\dag These authors contributed equally to this work.}
}

\begin{document}

\maketitle
\thispagestyle{empty}
\pagestyle{empty}

%%%%%%%%%%%%%%%%%%%%%%%    sections   %%%%%%%%%%%%%%%%%%%%%%%%%%%%%%%%%%%%%%%%%%%%%
% !TeX spellcheck = en_GB
\begin{abstract}

Adaptation to various scene configurations and object properties, stability and dexterity in robotic grasping manipulation is far from explored. This work presents an origami-based shape morphing fingertip design to actively tackle the grasping stability and dexterity problems. The proposed fingertip utilizes origami as its skeleton providing degrees of freedom at desired positions and motor-driven four-bar-linkages as its transmission components to achieve a compact size of the fingertip. 3 morphing types that are commonly observed and essential in robotic grasping are studied and validated with geometrical modeling. Experiments including grasping an object with convex point contact to pivot or do pinch grasping, grasped object reorientation, and enveloping grasping with concave fingertip surfaces are implemented to demonstrate the advantages of our fingertip compared to conventional parallel grippers. Multi-functionality on enhancing grasping stability and dexterity via active adaptation given different grasped objects and manipulation tasks are justified.
Video is available at \textbf{\tt\small youtu.be/jJoJ3xnDdVk/}.

\end{abstract}

% !TeX spellcheck = en_GB

\section{Introduction}

Human{\color{black}s} show fantastic adaptability when interacting with objects in environments, in both the sense of stability and dexterity, especially in grasping and manipulating objects. This capability can be attributed to {\color{black}a} swift sense of touch and muscle reflections, also to the powerful prior knowledge of grasping affordance for objects and ability to execution finger gaits adaptation before actual contact-making with the target object. In the society of robotics research, robustness and dexterity of robotic manipulation have been drawing intense attention throughout recent decades. Extensive stud{\color{black}ies} of grasping are presented and broadly defined as the method to immobilized objects w.r.t. the hand in \cite{mason2001mechanics} and the concept of force-closure and form-closure are brought to define the static balancing of grasped objects and immobilization of the grasped objects in the configuration space. These concepts are widely adopted and developed in {\color{black}the} following works. {\color{black}Dafle et al.} \cite{dafle2014extrinsic} extended to use motion cone to govern manipulation stability during in-hand manipulation while showing high dexterity with external force aiding the transformation of the object.  

Stability is usually used to address the asymptotic properties of a dynamic system. For robotics, since in most grasping tasks, fixing object relative to the hand is desired, stability of grasping systems {\color{black}is} critical. In a more microscopic view, a stable grasping system is equivalent to a system with contacts that can balance the object's weight and external forces, in another word, having equilibrium \cite{mason2001mechanics}. Traditionally, to increase conformity of the fingertip to objects' geometry and thus enhance grasping stability, rubber pads are attached to rigid fingers in industrial usage mimicking human finger pads' passive conformation. However, robotic systems do not have to be constrained by human's structure and capabilities, robot can surpass human in various perspectives, also in the grasping genre. For instance, {\color{black}Hou et al.} \cite{hou2019design} showed a gripper with stiffness variable fingertips featuring tunable fingertip hardness for situations require higher stability or precision. The fingertip proposed to conform to object passively, in comparison, in this work, our fingertip actively morphs given different manipulation tasks and object types. In terms of stability, our fingertip can form concave tip surface to provide a squeezing equilibrium that traps the object so as to secure a successful grasp.

\begin{figure}
	\centering
	\includegraphics[width=0.45\textwidth]{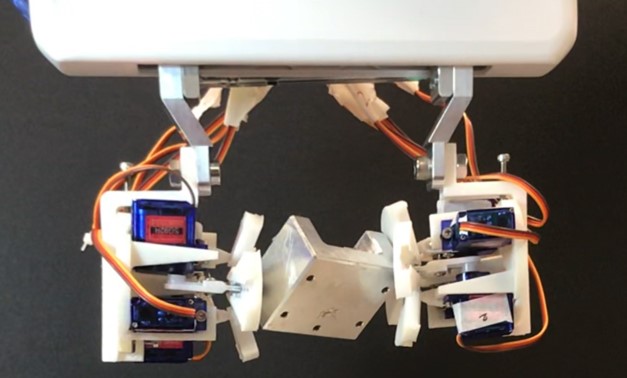}
% 	\vspace{-0.2cm}
	\caption{An example grasp using proposed morphing fingertips mounted on a parallel gripper with concave mode to form a power grasp. }
	\vspace{-0.5cm}
	\label{fig: general_grasp}
\end{figure}

Reorientation of the grasped object is often referred {\color{black}to} as in-hand manipulation. Dexterity of a robot hand is closely related to its capability to locate grasping sites accurately and perform in-hand reorientation maneuver. For precision grasp of small or delicate objects, pinching hand gaits are commonly seen in human daily. {\color{black}Borst et al.} \cite{borst2002calculating} studied how to calculate hand configurations to execute pinch grasping for robot hand. Besides, with a pinch gait, as the gripping force decreases, an object with its center of mass biased from grasping sites tends to rotate relatively to the hand, resulting in pivoting manipulation \cite{dafle2014extrinsic}. Pivoting involves {\color{black}a} reorientation of the grasped object without losing contacts w.r.t. fingertips, which is a convenient way to transform {\color{black}the in-hand} object
% in hand 
in 
% a 
chain operations of picking, reorienting, and placing/inserting. In our work, we design the fingertip that is able to morph to a convex surface with which a smaller contact patch can be made for the purposes of pinch grasp or pivoting manipulation. Furthermore, the proposed fingertip can perform tilting of the planar tip surface to reorient the grasped object while still maintaining grip.

This work contributes in the following folds:
\begin{itemize}
    \item Design a shape morphing enabled fingertip that shows 3 typical types of morphing primitives. The designed fingertip can switch between stable  and dexterous grasp modes as requested by different tasks. Validation of these types of morphing is implemented in multiple grasping tasks.
    \item Deriving the model of the morphing given types and degree of transformation, which is used as the inverse kinematics for the system.
\end{itemize}

The rest of this paper is organized as follows. Section \uppercase\expandafter{\romannumeral2} gives works related to state the research gap and advantages of our proposed design. Section \uppercase\expandafter{\romannumeral3} presents the design principle and describes the structure of the fingertip proposed. Morphing primitives that can be achieved by our fingertip are classified and their significances are stated. Section \uppercase\expandafter{\romannumeral4} derives the kinematics model. Section \uppercase\expandafter{\romannumeral5} validates the benefits claimed in grasping experiments. Finally, Summary is given and the conclusion is drawn in Section \uppercase\expandafter{\romannumeral6}.

% \subsection{Morphing primitives}

\section{Related Works}

Human{\color{black}s} use re-grasp, pose transition, and deformable adaptation on fingertip for the dexterity of grasping and manipulations targeting to distinct objects in the daily activities. Inspired from the bio-archetype, a number of studies are focusing on building an artificial hand that can mimic its skills. On one hand, employing re-grasp and pose transition for in-hand manipulation strategies has been widely studied from various aspects, showing great potentials in applications \cite{ma2011dexterity,antonova2017reinforcement,chavan2020planar}. On the other hand, the fingertip adaptation is still mainly developed and utilized for applications in the state of passive morphing via self-deformation of the soft material \cite{crooks2016fin,song2018fingertip,chavan2015two} or the flow characteristic of fluidic media \cite{brown2010universal,hou2019design}. Thus, while the passive morphing fingertip allows for limited dexterous in-hand manipulation \cite{hang2016hierarchical,sundaralingam2017relaxed}, an active morphing strategy is required for dexterous manipulations with more flexibility.

\subsection{Variable-frictional fingertip}

For the variety of properties on the fingertip, several researchers made efforts on the frictional change on the contact surface to alter the grasping performance. Becker et al. have proposed a friction switch mechanism for robotic hands via a pneumatic actuator \cite{becker2017tunable}. The adhesive-inflatable surface beneath the slippery plane can be extracted through holes installed on the slippery surface. The presented design is an extendedly active morphing strategy to the passive counterpart made by Spiers et al. \cite{spiers2018variable} which could switch to high-friction mode through force exerted in the finger. Other than friction variations on the solid pads, Mizushima et al. \cite{mizushima2019deformable} utilized lubricant to reduce friction coefficient for varying the contact environment. In addition, origami structure is also discovered for its morphing capability on the friction variation by shifting the inserted high friction surfaces using cable tension \cite{lu2020origami}.

\subsection{Cavity-shaped morphing fingertip for pivoting}

For processing fine manipulations and re-grasping tasks, multiple simple-structured two-phase hands rising extrinsic or intrinsic resources have been reported \cite{chavan2015two,chavan2018pneumatic,he2020soft}. They use mainly similar V-groove cavity structure on the parallel gripper to passively or actively reorient prismatic objects from a horizontal pose to an upright secured pose. When switching to upright secure mode, the soft materials employed, e.g. strip and silicon covering, will recede into the cavity, wrapping the cylinder and allowing contacts with the cavity. Thus an easy transition method between a firm grasp and a pinch grasp for free pivoting can be generated.

% \subsection{Passive morphing fingertip with fluid media}

% A vacuum-sealed package of granular beans is solid, yet becoming soft and flexible as relaxed. The granular jamming mechanism underlying can determine the morphology of the entire structure. From the bulky gripper \cite{brown2010universal} to the tiny fingertip \cite{hou2019design}, the morphing structures can adapt to target objects with a large range of dimensions. The adaptability depends on the bean size and the smoothness of the bean surface which affects the fluidic movement. However, the morphing strategy passively depends on the contact fit with the target.

In addition, Song et al. \cite{song2018fingertip} also put forward a collection of manipulation primitives for the representatives of contact features and three designed fingertips are evaluated {\color{black}on}
% for 
the effectiveness for a firm grasping facing different local surfaces. However, these primitives are manually fixed, which is lack of automation and not considerate for product transition of assembly line. Therefore, in order to extend the availability of the fingertip primitives, an active morphing strategy is proposed in this work with multiple morphing configurations for dexterous manipulations.

% !TeX spellcheck = en_GB
\section{Design Principle}

The goal of this work is to develop an active shape-morphing fingertip for adaptability to grasped objects' properties and task requirements, such as heavy weight, unstable holding for parallel flat grip{\color{black}p}ers,  pose adjustment and in-hand manipulation. Degree of freedom (DoF) of the fingertip shape morphing can be rather high \cite{follmer2013inform,hergenhan2014prototype}, even infinite \cite{becker2017tunable,song2018fingertip,he2020soft}. However, high DoF{\color{black}s} of the morphing could lead to under-actuation or {\color{black}a} high number of actuation, which hinder{\color{black}s} the further integration of the fingertip to grippers. To fully control the shape transition of the fingertip, full actuation is usually preferred. In this work, we adopt the thick origami \cite{chen2015origami} as the main skeleton of the morphing surface and use a four-bar linkage as the driving structure. As illustrated in Fig. \ref{fig: structure_illustration}, the fingertip has a{\color{black}n} origami structure with 4 leaf facets, 
% in 
which embed
% ded with 
guides and paired sliders to drive the flapping motion. This designed fingertip is self-contained with power and compact to benefit integration to conventional industrial parallel grippers as an easy improvement in terms of grasping adaptability.

% The GR2 gripper \cite{rojas2016gr2} illustrated the extended capabilities and benefits compared to the traditional two-finger gripper, with the effort of four-bar-linkage topology. The planar grasper can be easily controlled without any beforehand information on the target objects. Therefore, based on the principle of planar four-bar linkage, we have designed a shape morphing structure with 4 leaf facets constructed with 2 independent pairs of planar four-bar linkages. 
% % (
% Considering the predominant implementations of parallel grippers in industry, the morphing structure can be easily attached to the existing grippers and is aimed for fingertip adaptation.
% % )

\subsection{Structure design}

\begin{figure}
	\centering
	\includegraphics[width=0.5\textwidth]{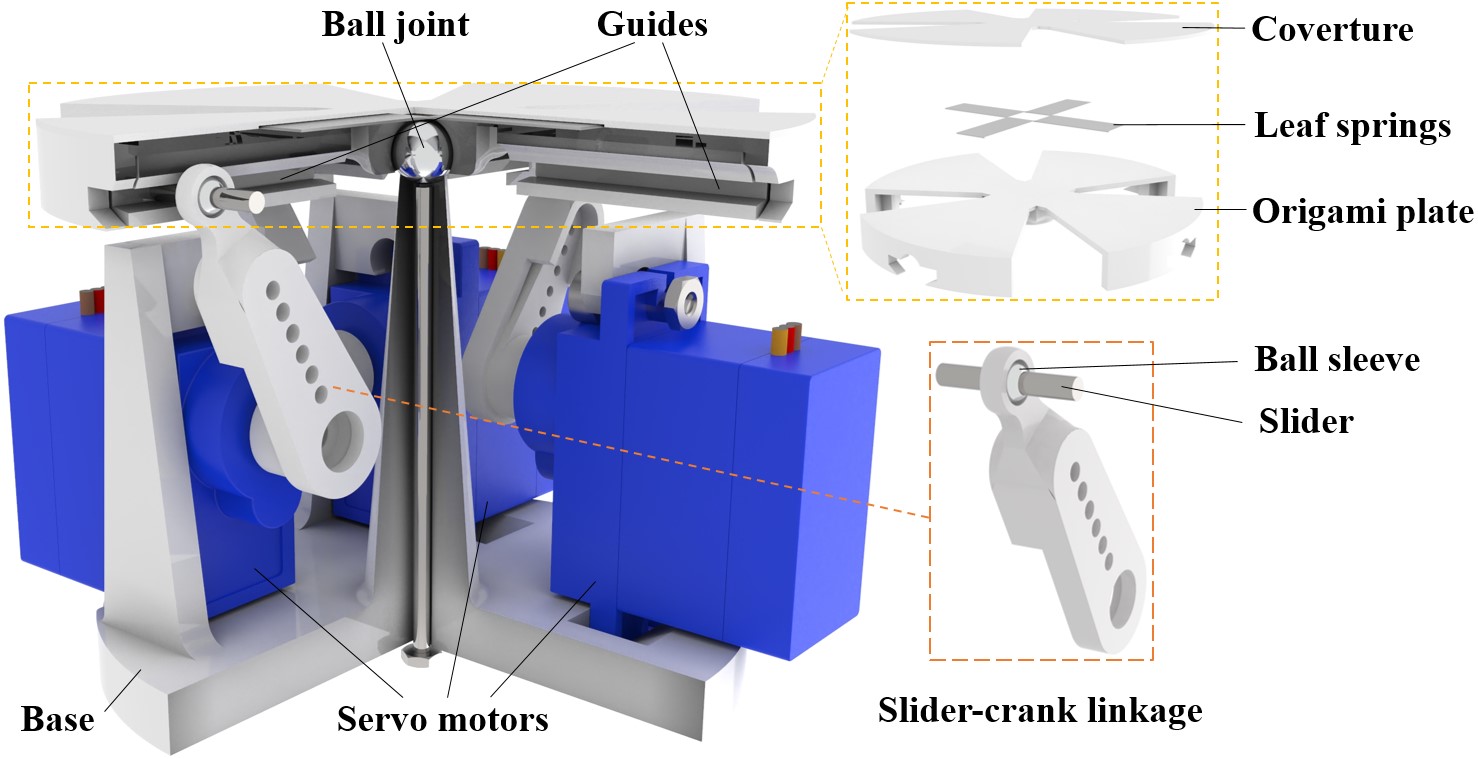}
	\vspace{-0.2cm}
	\caption{Illustration of a three-quarter section view of the morphing fingertip. The upper right inset indicates the detail {\color{black}of} the origami fingertip. The lower right inset shows the slider-crank linkage with a ball sleeve for free rotation. }
	\vspace{-0.4cm}
	\label{fig: structure_illustration}
\end{figure}

We design a morphing structure with a central terrace and four leaf facets, which is illustrated in Fig. \ref{fig: structure_illustration}. For the transformation between distinct grasping primitives, the transmission structure beneath the morphing surface is composed of two
% 4 
planar four-bar linkages which are mutually independent. Through the transmission mechanism, each leaf facet can be actuated individually by a servo motor. A ball joint is set at the center as a spatial pivot for the terrace when rotated in modes involves tilting.

Origami folding is an ancient technique to transform planar sheets into spatial structures 
% which 
{\color{black}that} can be implemented for multitudinous shape morphing transformations, and see its usage in locomotion and manipulations \cite{rus2018design,kan2019origami}. In our work, the complaint creases on a piece of morphing facet are seen as rotation shaft in order to provide additional constraints of the facets motion relative to the 
% center 
{\color{black}central} terrace and reduce the number of part{\color{black}s} and complexity of assembly. Meanwhile, it is noticeable that there are two redundant DoFs on the structure, one on each axis. In order to stabilize the under-actuated structure, we utilize leaf springs within the complaint rotational shafts for connections of arms and terrace. The main arm{\color{black}s} of the four-bar linkage structure are shown in Fig. \ref{fig: structure_illustration} and the cylindrical slider is inserted in the guide rail. Thus the slider will be restricted to translation along {\color{black}with} the guide and rotation relative to the servo arm extender. A ball sleeve is used to wrap around the slider for free rotation of the servo arm extender so that both four-bar linkages can move independently without coupled constraints.
With the translational and rotational movements of the slider within the guide and free rotation of leaf facet relative to the terrace, when the arm extender is driven by servo motor,
% about the spindle
 each leaf facet rotates by an angle related to the given angle of the paired servo motor through the mechanism of {\color{black}the} four-bar linkage.

\subsection{Morphing primitives}

Various movements of the morphing system are based on the motion of the 4 independent four-bar linkages, that is, slider-crank linkage (\textit{RRRP} type). The slider-crank mechanism used in the work is constructed by two cranks with a slider acting as the coupler in four-bar linkage which can slide along the crank and rotate the crank simultaneously. Via the actuation of servo arm extender, the leaf facet can be tuned upward and downward freely due to the existence of ball sleeves holding the cylindrical sliders.

With different combinations of rotation angles of the leaf facets, virtually infinite morphing configurations can be generated. Here, we select the following morphing primitives that are commonly encountered and important in robotic manipulation tasks.
% can be achieved to accommodate distinct local contact surfaces under various forms of manipulations.

\begin{figure}
	\centering
	\includegraphics[width=0.5\textwidth]{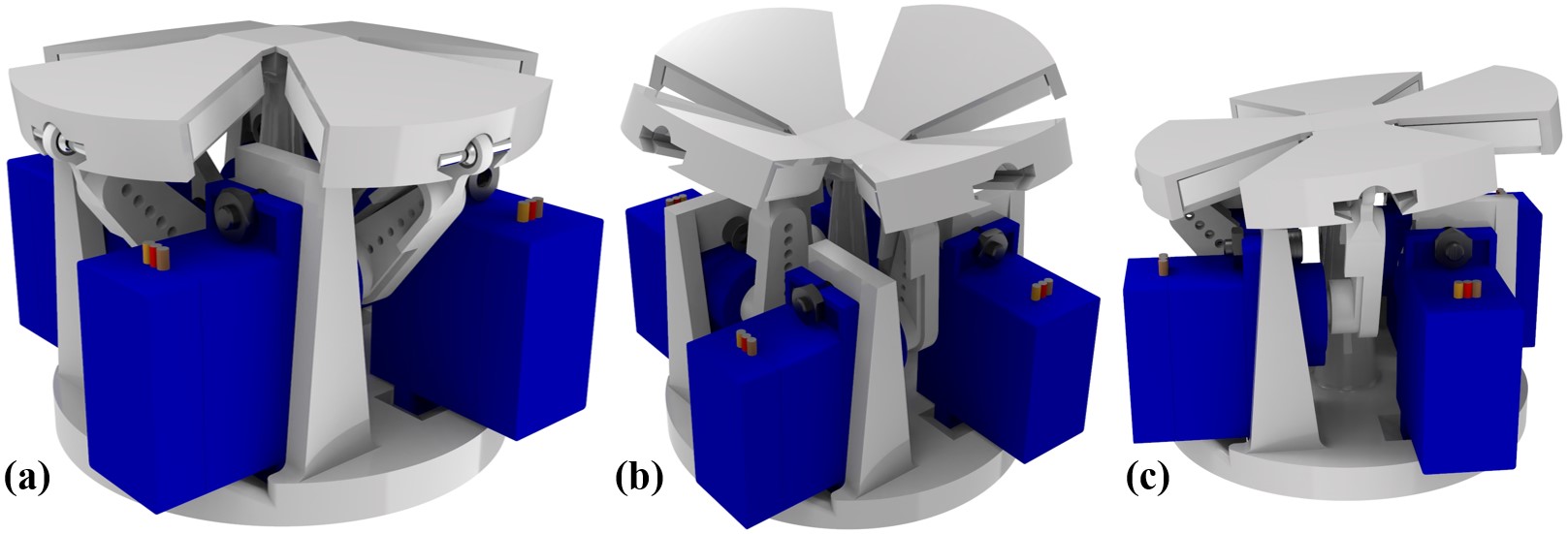}
	\vspace{-0.3cm}
	\caption{Rendered structure of the fingertip in three morphing modes. (a) convex mode. (b) concave mode. (c) tilted planar mode.}
	\label{fig:render3mode}
	\vspace{-0.4cm}
\end{figure}

\begin{enumerate}[i.]
	\item Firstly, as illustrated in Fig. \ref{fig:render3mode}(a), when four leaf facets rotate downwards simultaneously, the central terrace upheaves and a convex surface is generated. When this convex surface makes contact with {\color{black}an} object, the contact patch is relatively small, with which pivoting manipulation and pinch grasping can be performed. In pivoting, {\color{black}the} contact center is maintained while the relative rotation of the object w.r.t. the center is not constrained. This type of prehensile grasping is heavily studied in \cite{chavan2020planar}.
% 	Ideally a point contact is desired, however, it is more practically achievable to make contact with a small patch as Chavan et al. shows \cite{chavan2015two}.
    As the contact forces increase, the grasp can transit to pinch grasp, which fully constrains the object in hand. This type of grasp is useful when contacts at precise locations on objects are desired. These two types of grasp are sometimes combined sequentially to execute certain tasks and important to achieve more dexterous manipulation of objects.
% 	With a small surface area at contacts, there is always some degree of frictional resistance to counterbalance the gravitational torque and to transit the pivoting to a quasi-static pinch when a high-magnitude force is applied on an object with contact surfaces close to the center of gravity??. Although the high force generating a large friction is useful to firmly grasp an object, it is harmful to the gripper with high force concentration within specific area (and is unstable in practical implementations??), which is utilized for emergencies facing severe problems.

    \item On the contrary, if the leaf facets rotate with positive angles upwards (refer to Fig. \ref{fig:render3mode}(b)), the concave cavity will be formed. The concave cavity provides
    % a 
    stable local energy minimal to guide the grasped object to make contact with the terrace. By mimicking the contact geometries of the object surfaces when making contact, firm form-closure grasps can be secured.
    % Even though the fingertip cannot fit on the exact surface of the grasped object, the concave-shaped surface can conform to the geometry due to the redundant DoFs of the inserted springs.
    For objects that are challenging to grasp for conventional parallel grippers, such as object{\color{black}s} with sharp vertexes, large aspect ratios, this type of enveloping surface on fingertip can achieve better grasping stability, thus higher success rate.
    
    \item The third mode of the fingertip morphology goes to the tilting mode \ref{fig:render3mode}(c) when linkages rotate with angles with {\color{black}a} certain relationship that would be described in the modeling section. Naturally, with redundant DoFs on the terrace, the central platform can freely rotate. To eliminate the extra DoFs, leaf springs are installed on top of the flexible hinges between leaf facets and the terrace. Therefore, during each transition procedure, the terrace tends to move to the neutral state (the deflection of the leaf springs
    % are 
    {\color{black}is} minimal), in which the strain energy is minimized. 
    % In this mode, note that the tilting angle of the terrace can be arbitrarily defined by the rotational angles on the four servo motors, which extends the configuration space of the parallel gripper for surface fitting on the target object. 
    This mode can be used to perform object reorientation without releasing the grasping and in-hand manipulation if {\color{black}a} period of detaching is allowed in certain tasks.
    % Specially, with specific rotating angles on the opposite servo motors, the fingertip can form a tilted plane which is shown in Fig. \ref{fig:render3mode}(c).
\end{enumerate}

It is worthwhile noting that due to the existence of two redundant DoFs, the central terrace can move freely, even if the servo motors are fixed. 
% without stringent constraints around the ball joint in a balanced state with minimized strain energy. 
When free of contact, the terrace will rotate to an allowable pose within configuration space, approaching a posture whose normal can mirror the two opposite leaf facets. While an object is engaging the fingertip, the pose of the terrace will depend on the contact position and force exerted from the object that counterbalances the summation of the springs' deforming forces. Therefore, the fingertips have adaptability and tolerance to a certain amount of regulating residuals and grasping errors. The quasi-static
% modellings 
{\color{black}modeling} for typical morphing primitives is conducted in the following section.

\section{Kinematic Modeling and Primitive Transition Planner}

\begin{figure}
	\centering
	\includegraphics[width=0.495\textwidth]{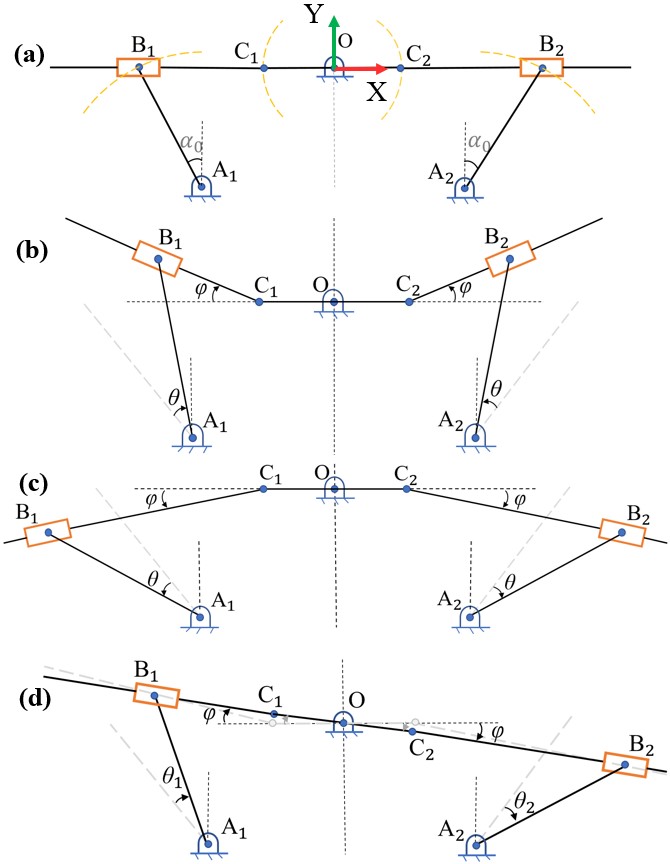}
% 	\vspace{-0.1cm}
	\caption{Illustrations of the simplified quasi-static model for the three typical morphing primitives. (a) Flat mode as the neutral state with the angle $\alpha = \alpha_0$ w.r.t. the plumb line. (b) Concave mode with two identical positive angles $\theta$ of the servo motors w.r.t. the neutral lines (in gray dash). (c) Convex mode with two identical negative angles $\theta$ of the servo motors w.r.t. the neutral lines (in gray dash). (d) successive quasi-static morphing sequences of tilted planar mode with one positive and one negative angle, $\theta_1$ and $\theta_2$, respectively.
% 	The yellow dashed curves indicate the possible motion trajectories of the corresponding points. The grey dashed lines suggest the positions of the linkages in the previous states.
	}
	\label{fig:model_3}
	\vspace{-0.5cm}
\end{figure}

In this section, the geometrical model of the fingertips would be derived. The model serves as an inverse kinematic formulation. With a chosen morphing primitive and a degree of the transformation, input angles of the servo motors can be computed with the model. 
As stated in \cite{balkcom2008robotic} from the perspective of 
% modelling
{\color{black}modeling}, though the paper bending is essential for paper-folding mechanics, it is of great help to simplify the model of {\color{black}the} structure as multiple rigid polygon facets with revolute joints connected. Therefore, in the letter, we introduce several assumptions as prerequisites. The assumptions are listed as follows.
% In order to reduce the complexity of the model, the following assumptions have been made:

\begin{enumerate}
\item The thickness of the leaf facets is neglected and the sliders move along the guides which intersect
% s 
with the ball joint.
\item All the linkage arms are considered as a rigid body and creases are seen as revolute joints whose width is neglected, without consideration of the strain energy during deformation which only exists within the leaf spring.
\item Only the modeling in a plane is studied due to {\color{black}the} decoupling of the structure in two orthogonal dimensions brought by the ball sleeves around the sliders. Motion modeling in two orthogonal dimensions is identical.
\end{enumerate}

In this work, we describe the morphologies of three typical configurations. Thus the 
% modelling 
{\color{black}modeling} mainly boils down to a geometrical relation between two inputs of rotating angles from servos and one output of tilting angle on the leaf facets. Specifically, for concave and convex problems, it is a one-to-one mapping due to two identical input angles. 

The initial position of the fingertip has a flat level with initial servo angles equals to $\alpha_{0}$, as shown in Fig. \ref{fig:model_3}(a). The servo angle is defined as the angle between linkage $AB$ and y-axis (plumb line). The configurations of the fingertips could be categorized mainly into 3 classes. Considering the symmetrical geometry w.r.t the centerline (vertical dash line) passing through $O$, the cases of concave and convex modes are schematically drawn in Fig. \ref{fig:model_3}(b-c). Let the angle between linkage $CB$ and x-axis (horizontal dash line) be $\varphi$, its positive value
% s 
indicates {\color{black}the} concave mode, and negative value denotes the convex mode. The concave mode is formed by rotating linkages $AB$ upward with positive $\theta$ w.r.t to the neutral lines, while in the convex mode, linkage $AB$ rotates in the opposite direction. In these two modes, the linkage $C_{1}C_{2}$ remains horizontal due to the symmetrical structure and actuation.
% s. 
Planar mode keeps the segmented linkages $B_{1}B_{2}$ in a straight line as shown in Fig. \ref{fig:model_3}(d), which means that servo motors' angles $\theta$ would follow certain relation to maintain the surface in the planar state. 

% \begin{figure}
% 	\centering
% 	\includegraphics[width=0.5\textwidth]{figure/model_graph_1.jpg}
% 	\vspace{-0.1cm}
% 	\caption{(a) The concave mode of the fingertip.(b) The convex mode of the fingertip.}

% 	\label{model_graph_1}
% 	\vspace{-0.3cm}
% \end{figure}
% \begin{figure}
% 	\centering
% 	\includegraphics[width=0.5\textwidth]{figure/model_graph_2.jpg}
% 	\vspace{-0.1cm}
% 	\caption{It is a step-by-step kinematic transformation of the planar mode from one angle to another. (a)The initial position of the planar mode. (Orange dot line indicates the possible motion trajectory of the linkage $C_{1}C{2}$.) (b)Linkage $AB_{1,2}$ rotate with angle $\theta_{1,2}$. (c) Linkage $C_{1}C_{2}$ rotates to form a straight line of $B_{1}B_{2}$, and thus, the planar rotates with angle $\varphi$.      }

% 	\label{model_graph_2}
% 	\vspace{-0.3cm}
% \end{figure}

In {\color{black}the} following, specific geometry relation between actuated angles and generated surface morphing types together with {\color{black}the} degree of transformations are derived.
The difference between concave and convex {\color{black}modes} depends on the range of $\theta$, and the transition between these two modes is continuous. In the cases of concave and convex modes, linkage $OC$ remains its initial pose, meanwhile, linkage $AB$ rotates by angles $\theta$ and sliders $B$ slide on linkages $CB$. The linkage $OC$ and $AB$ are represented in terms of vector in 
% eq.
(\ref{OC})-(\ref{AB}) {\color{black}respectively}. The Cartesian coordinate frame is defined in Fig. \ref{fig:model_3}(a). 

\vspace{-0.1cm}
\begin{equation}
    \overrightarrow{OC}=\begin{bmatrix} OC_{x} \\ OC_{y}\end{bmatrix}=\begin{bmatrix} \pm l_{oc}\\ 0\end{bmatrix}\\
    \label{OC}
\end{equation}
\begin{equation}
    \overrightarrow{AB}=\begin{bmatrix} AB_{x} \\ AB_{y}\end{bmatrix}=\begin{bmatrix}\pm l_{ab} \;\sin{(\alpha_{0} \pm \theta)} \\ \pm l_{ab} \; \cos{(\alpha_{0} \pm \theta)}\end{bmatrix}
    \label{AB}
\end{equation}
where $l_{oc}$ is the length of the linkage $OC$, and $l_{ab}$ is the length of linkage $AB$.

The relative position between origin and point $A$ is the design parameter. Therefore, $\overrightarrow{OA}$ is known. $\overrightarrow{AC}$ and $\overrightarrow{CB}$ could be calculated following 
% eq.
(\ref{CB}).

\vspace{-0.3cm}
\begin{equation}
\begin{split}
    \overrightarrow{AC}&=\overrightarrow{OC}-\overrightarrow{OA}\\
    \overrightarrow{CB}&=\overrightarrow{AB}-\overrightarrow{AC}\\
    &=\begin{bmatrix}OA_{x}+ AB_{x}-OC_{x}\\OA_{y}+ AB_{y}\end{bmatrix}
\end{split}
\label{CB}
\end{equation}

The trigonometry relation of $\overrightarrow{CB}$ is listed in 
% eq.
(\ref{phi_1}). By solving the equation given $\varphi$, the input angle of the servo is deduced for concave and convex modes. The $\varphi$ is negatively proportional to $\theta$.

\vspace{-0.3cm}
\begin{equation}
    \varphi = \arctan (CB_{x},CB_{y})
    \label{phi_1}
\end{equation}

For the planar mode, the linkages $OB$ rotates by an angle $\pm \varphi$ on each side to make sure the structure maintains the planar configuration. The vector of linkages $OB$ is in 
% eq.
(\ref{OC_2}).

\vspace{-0.3cm}
\begin{equation}
\begin{split}
    \overrightarrow{OB}_{1,2}&=\begin{bmatrix}\pm l_{ob} \cos(\pm\varphi) \\\pm l_{ob} \sin(\pm\varphi)\end{bmatrix}\\
    % =\overrightarrow{OC}+\overrightarrow{CB}\\
    &=\overrightarrow{OA}+\overrightarrow{AB}
    =\begin{bmatrix}OA_{x}+ AB_{x}\\OA_{y}+ AB_{y}\end{bmatrix}
    \end{split}
    \label{OC_2}
\end{equation}

The geometry relation
% s 
in planar mode is the same as concave and convex modes, 
% so that eq.
{\color{black}thus} (\ref{AB})-(\ref{CB})
% are 
remain true. {\color{black}The equation solving} 
% Solving 
the servo input angle on each side is shown in 
% eq.
(\ref{phi_2}).

\vspace{-0.4cm}
\begin{equation}
    \pm \varphi = \arctan (OB_{1,2\;x},OB_{1,2\;y})
    \label{phi_2}
\end{equation}

\section{Experiments and Results}

In this section, two performance tests and manipulation demonstration{\color{black}s} have been conducted to characterize the transformability of the proposed fingertip and evaluate the efficiency of the grasping planner derived from 
% modelling 
{\color{black}modeling} in the previous section. First, the dimensions, fabrication details, and experimental platform setup for the fingertip are clearly stated. 
% Next
{\color{black}Moreover}, {\color{black}an} angular stroke test is proceeded for verification of the prediction of proposed modelling and planner. Additionally, a configuration space test for planar tilting mode is conducted to show the convenience for in-hand manipulations. Finally, a grasping demonstrat{\color{black}ion}
% e 
on multiple parts with typical surfaces and shapes is proceeded to present the practical performance of the morphing fingertip.

\subsection{Prototyping and experimental setup}

Although nowadays traditional CNC machining method{\color{black}s} for mechanical removal of materials are prevalent in the automation system, additive manufacturing technique provides a novel fabrication method and forming process, especially for parts with complex geometry shapes and robots with numerical tiny parts and complicated assembly. 
% The direct 3D printing manufacturing shows multiple advantages, e.g. rapid processing to shorten the iteration period between modelling and prototyping, ease of design without considerations on tooling and multistage processes, simple control on the stiffness of parts via infill of the structure, and improvement of integrity on the entire structure \cite{gibson2014additive}.
In this work, based on previous experiences \cite{kan2019origami}, rigid and soft materials are utilized with 3D-printing method and off-the-shelf 3D printers for multiple parts of the fingertip. 
The base for support and the guides for sliding are fabricated by stereolithography (SLA)
% % additive manufacturing method 
% in Formlabs 3D printer (Form 2). 
% % , Formlabs
% % with standard resin 
% SLA-printed parts are 100\% infilled and can thus hold a large payload, which is suitable for parts under complicated loading circumstances. 
The origami plate for soft fingertip is printed by a fused-deposition-modeling (FDM) 3D printer (Ultimaker 2+, Ultimaker) with soft thermoplastic polyurethane (TPU) filament. 
% Owing to the soft material, patch contact with small contact area can be generated on the leaf facets with target object for better positioning and holding. 
For the other linkage parts, the same FDM machine is employed with rigid PLA filament.
% , on account of the lightweight property of the material compared to resin, reducing the inertia of the part during movement.
The dimensions of the origami plate are with {\color{black}a} diameter of 55mm and thickness of 1mm, connected with flexible hinges of 10mm long, 2mm wid{\color{black}e}
% th 
and 0.6mm thick.
% , based on empirical experiences after several experimental testing. 
% The stiffness of the soft fingertip can be tuned with various hardness of TPU and diverse infill percentage. For the sake of requirements of robustness in practical grasping manipulations, in this prototype, we use a TPU with shore hardness of 95A and the infill percentage is set to be 100\%. 
In addition, for the parts with continuous movement, metal parts with frictionless feature are considered for smoothness of motion, e.g. ball joint, sliders, and ball sleeves. 
% Meanwhile, steel with a large range of linear elasticity is inserted within the flexible fingertip as leaf springs for balance and stabilization of the entire structure. 

For verifications and characterizations on manipulation performances, we attach the morphing fingertips on both fingers of the parallel gripper on a robotic manipulator (Panda, Franka Emika) for multiple characterization tests and grasping demonstrations. The robotic arm acts as a platform to localize the target object and position the parallel gripper.
% , with its control repeatability of 0.1mm. 
% For actuation sources, each soft fingertip prototype is actuated by four low-cost servo motors (Tower Pro SG92R, 1.7USD each), mounted on the base in an equally distributed manner, with each driving a leaf facet individually through four-bar linkages. 
% The servo motor is in small dimensions, which provides the possibility of constructing a compact structure.
The servo motors (Tower Pro SG92R) are connected to a controller board 
% (Ardruino 
({\color{black}Arduino} Mega microcontroller) to switch the input signals manually and transit within various modes. A connector
% fixture 
is also arranged for linking of the fingertip with existing parallel gripper. 
% For the tilting configuration space test, a stick is installed above the fingertip, acting as a pointer and representing the normal direction.

\subsection{Angular stroke test in convex and concave modes}

\begin{figure}
	\centering
 	\includegraphics[width=0.4\textwidth]{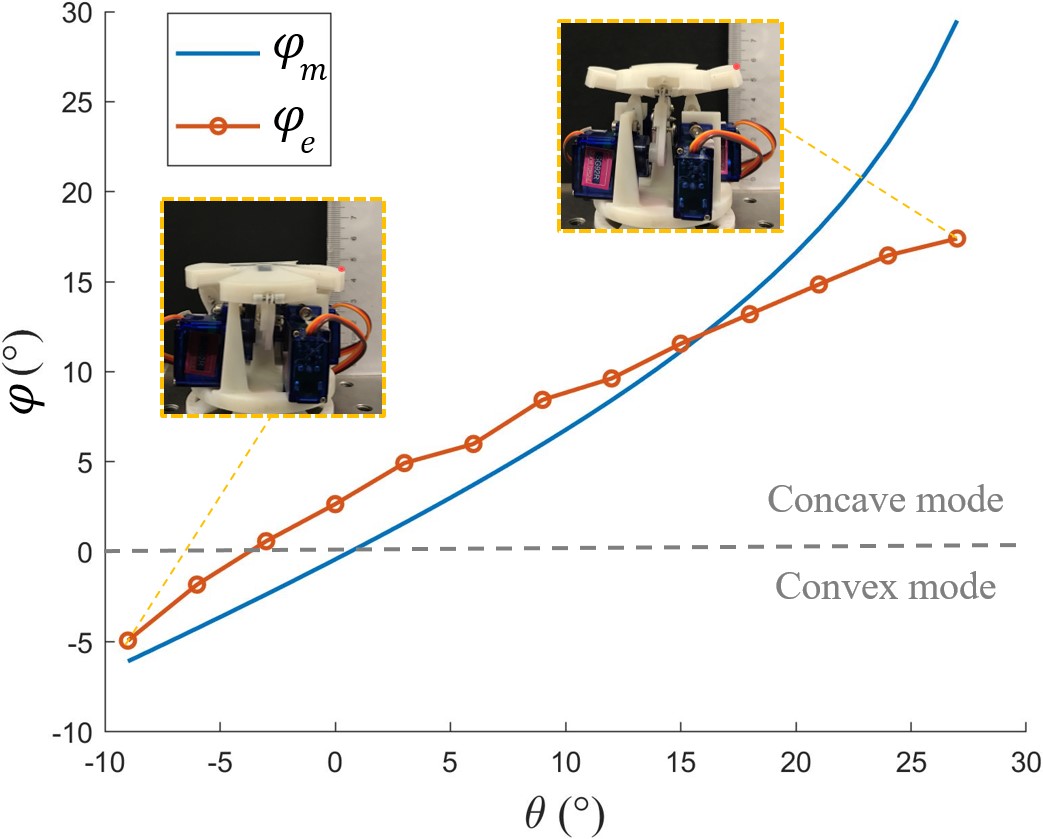}
 	\vspace{-0.2cm} 	
%  	\vspace{-0.3cm}
	\caption{Comparison of model {\color{black}prediction}
% 	planner 
	and the practical experimental results in the convex and concave modes, given an input rotational angle from the servo motor. The blue curve and the orange curve with scatters represent the output angles of model prediction and the experiment data respectively. The concave and convex modes are separated by $\varphi=0$. The insets show the measurement states at the start and end points.}
	\label{fig:angularstroke}
	\vspace{-0.4cm}
\end{figure}

In this test, the position shift of the leaf facet is investigated upon four-bar linkage transmission driven with discrete increasing magnitude of rotation angle on the servo motor which corresponds to angle $\theta$ stated in Section \uppercase\expandafter{\romannumeral4}. The actuation procedure is separated into 13 sequential steps, each of which is a three-degree rotational input, with three-second delay intervals for measurement in a stationary state. For the versatility of the test results and ease of comparison with the model predictions, in this study, the controlled angle on the servo is designated as the input variable while the calculated tilting angle of the leaf facet serves as the output variable derived from the measured height of the facet tip. The measurement has been repeated on four leaf facets for three trials each with the average utilized for characterization. A camera is placed in front to record the position swing of the red marker attached to the facet tip in the procedure and the test results are shown in Fig. \ref{fig:angularstroke}.

In Fig. \ref{fig:angularstroke}, an angular mapping prediction curve of $\theta$ vs. $\varphi$ from {\color{black}the} model is presented. Multiple discrete points are captured from the real experiment for practical mapping relationship with insets of morphing states at the key points. At the start point, the slider moves approaching the proximal limit of the guide and the leaf facet is pushed upwards in the peak position. Afterward,
% s, 
it shows that under the actuation of servo motor, the facet moves downwards, converting from concave into convex mode. During the 13 sequential steps, the servo motor rotates 12 times. The data indicate the rotation angle of the leaf facet in the concave and convex modes proceeds a shrinking turn with a decreasing scale factor from input to output as the arm of servo arm lowers. 
% (, rather than linearly correlated relationship with the input) whose relationship can be fitted with polynomial least square (LS) method. Here we choose the cubic LS curve which is illustrated in Fig.XX. the scale factor is easily acquired to be XX at the starting point, while reducing to XX at the low end. However, the results indicate a gap between model prediction and practical experiment data. It is explainable that the difference exists on the prototype, resulting from the assumptions in the modelling, fabrication tolerance and the performance instability and inconformity of the servo motors. It can be expected that with more precise CNC machining and more stable actuators, the error can be mitigated in the future revisions.
In Fig. \ref{fig:angularstroke}, the results indicate an approximately constant gap between model prediction and practical experiment data in the convex mode, probably due to the simplification in the {\color{black}modeling}
% modelling 
and the performance instability and inconformity of the servo motors. After converting into concave mode with a large rotational angle on the leaf facet, the gap rises 
% rapidly. 
{\color{black}continuously}. It is explainable that the difference exists on the prototype, resulting from the fabrication tolerance, especially the fit tolerance in the ball joint. When the leaf facets are pulled up, the soft cover on the monolithic origami plate outside the ball joint will tend to break away from the closure, which significantly remits the angular deformation. Therefore, it can be expected that with more precise CNC machining and more stable actuators, the error can be mitigated in the future revisions.

\subsection{Configuration space test of plane morphing fingertip}

As mentioned in Section \uppercase\expandafter{\romannumeral3}, the central terrace can rotate w.r.t. the surrounding leaf facets. Thanks to the leaf springs for force balance and geometry constraints, the configurations of the fingertip tend to a neutral state with minimal strain energy within the deformable springs. Especially, with specific rotating angles on the opposite servo motors, the morphing surfaces can form a tilted plane, that is, tilted planar mode. In order to characterize the tilting space of this configuration, a golden rod with a metal top is attached to the terrace as a pointer directing to the normal orientation. It is obvious that with distinct drive combinations from the four leaf facet, the top of the central rod is steered to move approximately in a two-dimensional plane. By tilting the rotating angles on the two pairs of servo motors, a number of positions can be attained by the rod with sequential actuation transitions between adjacent poses.

\begin{figure}
	\centering
 	\includegraphics[width=0.45\textwidth]{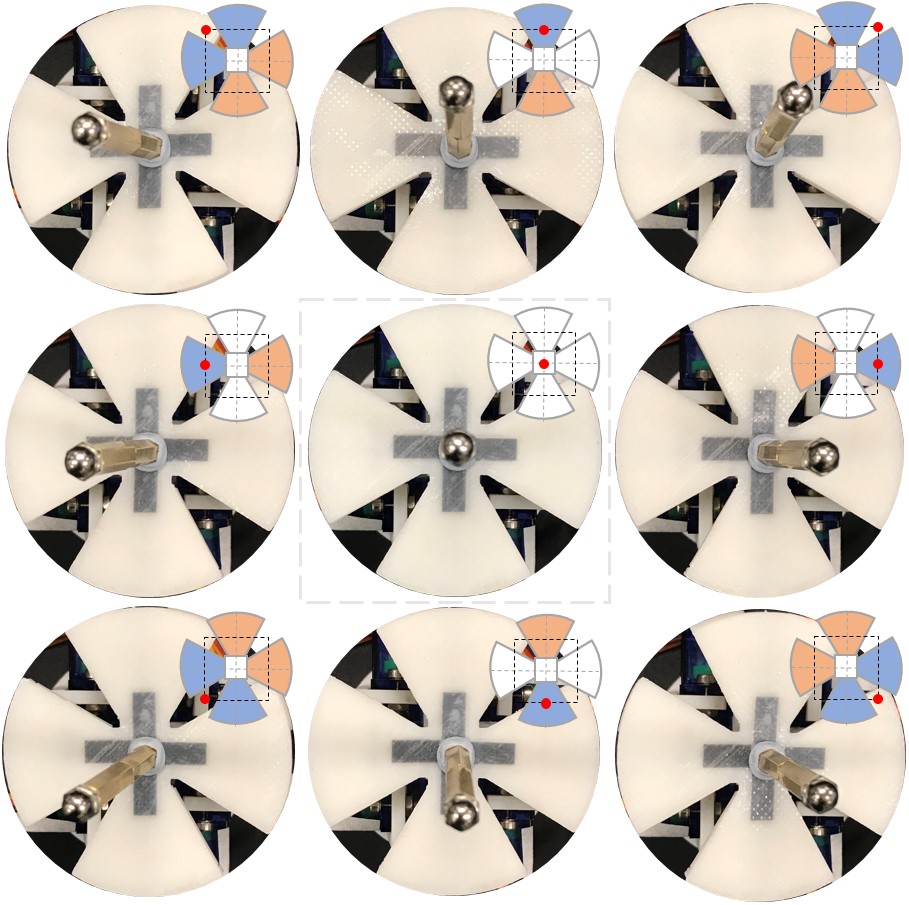}
	\caption{Eight distinct poses within the configuration space of the planar morphing fingertip by exerting combinations of actuation to servo motors. A neutral position is shown in the middle when no rotation angle is applied with the plane laid flat horizontally. The blue and orange leaf facets represent configurations with a negative (downward) and positive (upward) angle respectively. The red dot denotes the current position of the pointer top and the black dashed square indicates the theoretical trajectory of the pointer top under sequential actuation.}
	\label{fig:planar8dir}
	\vspace{-0.4cm}
\end{figure}

Theoretically, through successively tuning the rotating angles on the two pairs of cranks in the four-bar linkages, the trajectory of the pointer top can generate a square shape which refers to the black dashed square in Fig. \ref{fig:planar8dir}. Here, we illustrate eight typical positions on the trajectory with corresponding actuation states in Fig. \ref{fig:planar8dir}. The blue leaf denotes pose with a negative rotation angle downwards and the orange leaf denotes pose with a positive and upward one, while the white facet represents a relaxed state in which the surface is coplanar with terrace and free of rotation following the constraints from current actuation. The red dot shows the current position of the pointer top. The results in Fig. \ref{fig:planar8dir} indicate the varieties of the tilting configurations with different driving modes. The flexibility on the fingertip pose provides conveniences for in-hand manipulations. The details of verification on the practical grasping are described in the next section.

\subsection{Manipulation demonstration}

To explore the performance potential and verify the proposed design of the soft morphing fingertip, a wide range of objects are selected for manipulations.
% as summarized in Fig.XX. 
The pile of target objects with distinctive surface types is suitable for different manipulation tasks. The graspers without consideration on the fingertip adaptations have {\color{black}a} limited scope of target objects on gripping \cite{kan2019origami}, especially on weight. Utilizing the proposed fingertip, it can efficiently improve the grasping capability on payload and adaptation of features, rather than making use of the deformation of the material within the contact area for better accommodation.

\begin{figure}
	\centering
 	\includegraphics[width=0.45\textwidth]{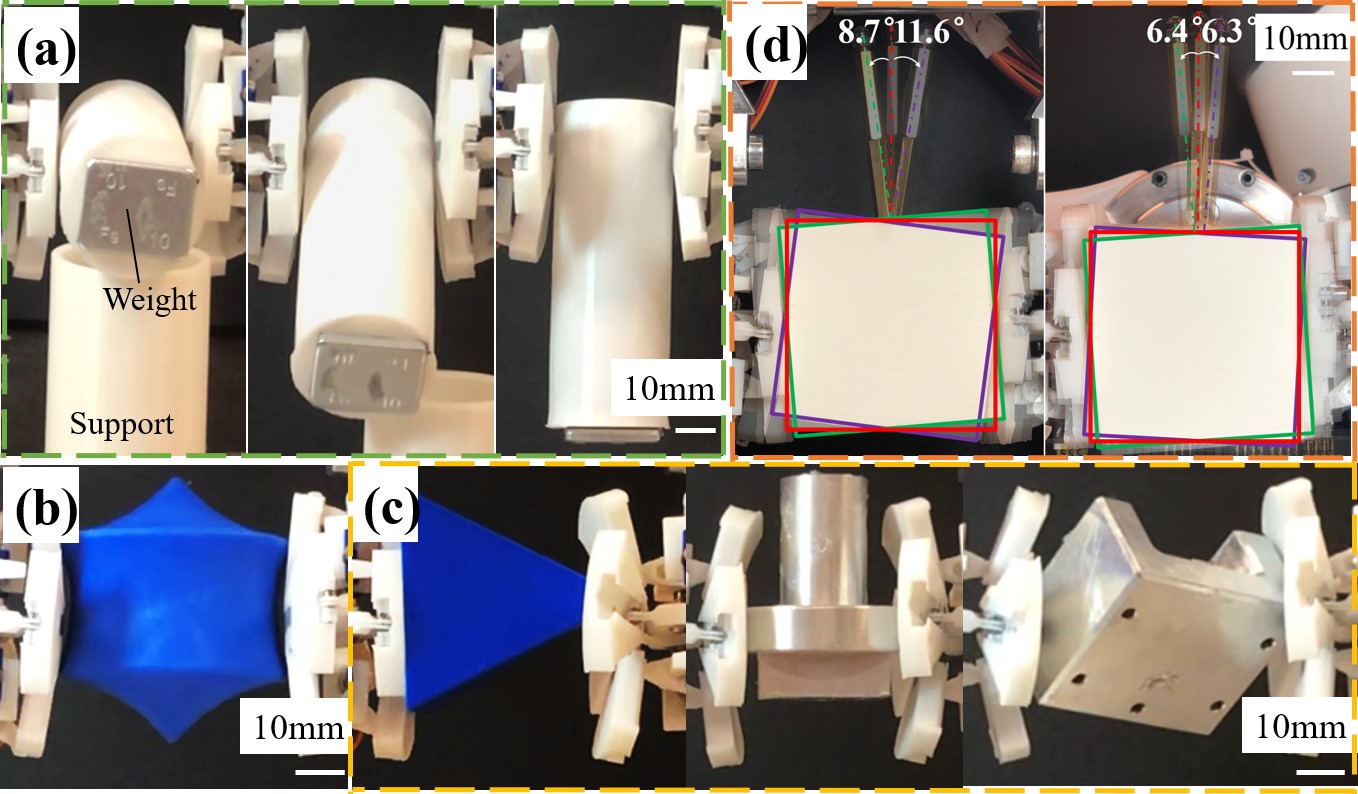}
	\caption{Manipulation of distinct types of objects in different modes. (a) pivoting process sequences in the convex mode. (b) pinch in the convex mode. (c) power grasps in the concave mode. (d) reorientation in-hand manipulation in the tilted planar mode.}
	\label{fig:graspdemo}
	\vspace{-0.4cm}
\end{figure}

\subsubsection{Pivoting with convex mode}

Pivoting is a significant re-grasping strategy for in-hand manipulation, instead of releasing down and picking up again. For these dexterous tasks, robotic hands with multiple fingers are often essential to complete a successful strategy \cite{antonova2017reinforcement}. With the prevalence of parallel grippers with flat fingertips employed in the industry, however, it is important to develop a morphing fingertip alternative with the capability of pivoting without much additional actuation complexity. To validate the performance effectiveness, here we use a typical cylinder which is identified as predominant parts in the industry manufacturing \cite{groover1986industrial,chen1982gripping}. The 3D-printed cylinder is light-weighted, thus for the ease of accomplishment of the pivoting task with self-gravity, the torque from gravity should be augmented by a weight attached at the end tip. 

From Fig. \ref{fig:graspdemo}(a), it can be noticed that the fingertips manipulate at the other end tip. After the morphing transition, the contact between the cylindrical part and the convex-shaped fingertips remains a small-patch contact during the pivoting phase. With the torque applied on the object as a result of gravitational force from the weight, pivoting can be achieved in the grasp. In addition, it should be mentioned that the small patch contact still generates little resistance to the pivoting procedure, for which the weight attached is essential in this task.

\subsubsection{Pinch with convex mode}

As stated in Section \uppercase\expandafter{\romannumeral3}, pivoting along is not sufficient for {\color{black}the} reorientation of objects. The gripper should have capability to switch between pivoting and a firm grasp, that is, pinch. The transition can be accomplished via deformation of the fingertip, as shown in \cite{chavan2018pneumatic}. Traditionally for the parallel grippers, when facing an object with {\color{black}a} concave surface, this area will not be an ideal position for contact owing to the geometrical mismatch. Instead, it will explore another pose for better interactions. Some soft grippers are capable {\color{black}of}
% to 
this grasping task via passive morphing of the soft materials \cite{crooks2016fin} or fluidic media \cite{brown2010universal}. By exerting forces on the object through the soft material, the material can deform to fit the local contact surface. In our design, we combine the soft structure with {\color{black}an} active morphing strategy to dynamically match the contact profile. An irregular spindle-shaped 3D-printed part is utilized to validate the feasibility of the pinch grasp.

Fig. \ref{fig:graspdemo}(b) indicates the grasping pose for {\color{black}a} pinch of concave-shaped object. A convex-shaped profile is formed on the morphing fingertip which pairs in the position of concave contour on the object and secure{\color{black}s} the holding to a pinch grasp. Since the geometric adaptation and constraints on the contact range, a large squeeze force for frictional holding is not necessary.

\subsubsection{Power grasp with concave mode}

The concave-shape object can be handled by a gripper with {\color{black}a} convex-shape fingertip, while the object with {\color{black}a} convex outline, on the contrary, can be lifted using a concave pairing surface. Generally, wrapping up the target object by means of {\color{black}a} concave-shaped surface is an intuitive method for a stable grasp, especially with applications on multi-fingered gripper for a form-closure, even force-closure, grip. Therefore, the concave mode of the proposed fingertip can be applied to a wide range of objects under different scenarios. Utilizing the geometric limitations for kinematic constraints, the gripper in the concave mode can tolerate little grasping error. The groove cavity has capability to re-align and localize a mismatched object to a proper pose and later secure the grasp and maintain the pose, following the gripper to move around freely.

In this work, a number of target objects are listed for the applicability of the morphology of the fingertip. The selected objects with special shape features and grasping poses are challenging tasks for traditionally parallel grippers, e.g. sphere/ellipsoid-shaped objects, pentahedron, half-Oloid roller, and body-diagonal grasping on the cube. When facing these target objects, the fingertip can morph to
% a 
proper concave shape for better contact adaptation on the local surface. The details can be obtained from Fig. \ref{fig:graspdemo}(c).
% and the attached video. 
Considering the lightweight properties of the grasped objects stemming from the manufacturing technique, two additional metal parts with block and cylindrical profiles are grasped for practical evaluations on industrial applications. In order to differentiate the grasping performance of the gripper with parallel fingertips and concave-shaped fingertips, we try to shake the robotic arm during the handling process. It is worthwhile noticing that the morphed fingertips can guarantee stable grasps and successful lift-up, while the parallel gripper with the same grasping gap cannot finish the entire operation, even failing on holding and lifting at the starting point. The multiple tests verify the effectiveness and applicability of the soft morphing fingertips for practical parallel grippers.

\subsubsection{In-hand manipulation with tilted planar mode}

In-hand manipulation is an ability to relocate and reorient the object within hand which can be classified into two types, extrinsic and intrinsic dexterities according to the actuation resources for manipulations \cite{dafle2014extrinsic}. The intrinsic dexterity manipulations require the grippers with higher DoFs than that of the target objects \cite{dafle2014extrinsic}. However, since the DoFs of the entire system is equal to that of {\color{black}a} movable object in SO(3), it is theoretically impossible to fulfill the in-hand manipulation tasks using the morphing fingertip. With the fabrication residuals, the structure accomplishes in-hand manipulations for {\color{black}the} reorientation of a cube within {\color{black}a} specific operation space. In Fig. \ref{fig:graspdemo}(d), based on the configuration space of planar mode abovementioned, the gripper can rotate the block in roll axis for $-11.6^\circ$ to $+8.7^\circ$ and in yaw axis for $-6.3^\circ$ to $+6.4^\circ$. It is noticeable that due to the over-constrained configuration and gravitational force during manipulation, the target objects for tilting will slide along contact surfaces, tending downwards to be squeezed out of working place. Despite {\color{black}the} lack of a large scale of freedom, the morphing fingertip still provides a range of flexibility for the intrinsic in-hand manipulations of cubic object{\color{black}s}.

% \begin{enumerate}[i.]
% 	\item 
	
% \end{enumerate}

% !TeX spellcheck = en_GB
\section{Conclusion and Future Works}
In this letter, we have presented a novel under-actuated morphing fingertip utilizing origami panels, four-bar linkages, and servo motor actuation. The four leaf facets around the central terrace can be controlled independently via the rotation of the servo arm and the transmission of the four-bar linkage. With multiple DoFs operated, the morphing structure can be transmitted into three typical surface primitives which are of great benefit to dexterous grasping and in-hand manipulations. A kinematic model has been proposed for the three grasping modes through geometric relationship{\color{black}s} and constraints. Therefore, a manipulation planner {\color{black}can be} 
% is 
derived from the geometric model. Soft morphing fingertips have been prototyped through multiple 3D printing techniques with soft and rigid materials, working as an alternative of flat fingertips on the parallel gripper. Through experiments, the fingertip is validated with {\color{black}an} angular stroke test for {\color{black}the} indication of ease of control and the property of reorientation for dexterous manipulations. This letter shows the potential to apply the morphing fingertip on the existing parallel grippers prevalent in the industry. The experimental demonstration shows that the soft fingertip is able to transit into multiple morphing primitive and improve the dexterity of parallel gripper to handle a wide range of objects with distinct surface features and to extend the manipulation capability in real applications.

However, some limitations and challenges still remain to be problems. Due to the fabrication tolerances and the performance instability and inconformity of the servo motors,
% ., 
the practical performance cannot perfectly match the model prediction. In addition, the simplification of the assumptions in the model can be another cause for the data mismatch and the changes in the geometric relationship generate an approximately constant offset. Therefore, in the next step, the structure should be manufactured by CNC machining for smaller fabrication tolerances, finer transmission resistance, and higher control accuracy, and be employed with a smaller and more stable servo motor for miniaturization of the system. Meanwhile, the model needs to be re-established without considerations on many current assumptions for more precise predictions on motion trajectory and grasping planner. Lastly, tactile or vision-based sensors can be added for a closed-loop interaction with contact match.

%\input{acknowledgment}
%%%%%%%%%%%%%%%%%%%%%%%%%%%%%%%%%%%%%%%%%%%%%%%%%%%%%%%%%%%%%%%%%%%%%%%%%%%%%%%%%%%

%\input{CASE2020.bbl}
\bibliographystyle{IEEEtran}
\bibliography{CASE2020}

\end{document}